\documentclass[runningheads]{llncs}
 
\usepackage{eccv}
\usepackage{makecell}

\usepackage{eccvabbrv}

\usepackage{graphicx}
\usepackage{booktabs}

\usepackage[accsupp]{axessibility}  
\usepackage{float}

\usepackage{hyperref}

\usepackage{orcidlink}
\def\methodname{SynthSet} 
\begin{document}

\title{\methodname: Generative Diffusion Model for Semantic Segmentation in Precision Agriculture} 

\titlerunning{\methodname: Generative Diffusion Model for Semantic Segmentation}

\author{Andrew Heschl\inst{1}\orcidlink{0009-0008-1400-9897} \and
Mauricio Murillo\inst{1}\orcidlink{0009-0003-8650-4292} \and
Keyhan Najafian\inst{2}\orcidlink{0000-0003-3050-8107} \and \\
Farhad Maleki\inst{1}\orcidlink{0000-0002-5673-8210}
}

\authorrunning{Heschl et al.}

\institute{University of Calgary, Calgary, Alberta, Canada \\
\email{\{andrew.heschl, mauricio.murillo, farhad.maleki1\}@ucalgary.ca} 
\and
University of Saskatchewan, Saskatoon, Saskatchewan, Canada \\
\email{keyhan.najafian@usask.ca}
}

\maketitle

\begin{abstract}
This paper introduces a methodology for generating synthetic annotated data to address data scarcity in semantic segmentation tasks within the precision agriculture domain. Utilizing Denoising Diffusion Probabilistic Models (DDPMs) and Generative Adversarial Networks (GANs), we propose a dual diffusion model architecture for synthesizing realistic annotated agricultural data, without any human intervention.
We employ super-resolution to enhance the phenotypic characteristics of the synthesized images and their coherence with the corresponding generated masks.
We showcase the utility of the proposed method for wheat head segmentation.
The high quality of synthesized data underscores the effectiveness of the proposed methodology in generating image-mask pairs.
Furthermore, models trained on our generated data exhibit promising performance when tested on an external, diverse dataset of real wheat fields. The results show the efficacy of the proposed methodology for addressing data scarcity for semantic segmentation tasks. Moreover, the proposed approach can be readily adapted for various segmentation tasks in precision agriculture and beyond.
\end{abstract}
%
\section{Introduction}\label{sec:intro}
In recent years, deep learning has played a pivotal role in precision agriculture~\cite{Marinello2023ThePT}, particularly for tasks such as object detection~\cite{Khaki2021WheatNetAL} and image segmentation~\cite{Li2023EnhancingAI}, owing to its ability to extract valuable information with unprecedented accuracy and efficiency. Though deep learning is a powerful tool, in many domains, data-related challenges hinder the development of robust models, particularly in tasks where pixel-level data annotation is required~\cite{Najafian2022SemiSelfSupervisedLF}. Manual annotation is time-consuming, costly, and subject to inconsistencies and errors. Therefore, the challenge of obtaining large-scale annotated datasets for supervised learning in tasks requiring pixel-level annotations, such as semantic segmentation, remains a bottleneck for developing generalizable models. To address this, methods for synthetic data generation have gained traction, providing an alternative for developing extensive labeled datasets with minimal human intervention~\cite{Najafian2022SemiSelfSupervisedLF}. \par 
The field of computer vision also has witnessed remarkable advancements, particularly in the area of generative models. Among these, Denoising Diffusion Probabilistic Models (DDPMs) have shown promising performance in generating high-quality images, outperforming methods~\cite{Dhariwal2021DiffusionMB} such as Generative Adversarial Networks (GANs)~\cite{Goodfellow2022GenerativeAN} and Variational Autoencoders~\cite{Kingma2019AnIT}. DDPMs can produce diverse images and are less prone to issues such as mode collapse during training. DDPMs are suitable for various tasks, including in-painting~\cite{lugmayr2022repaintinpaintingusingdenoising}, super-resolution~\cite{rombach2022highresolutionimagesynthesislatent}, and text-conditioned~\cite{latentdiffusion} image generation. \par
To address the data annotation bottleneck, we propose a novel DDPM-based architecture for generating pixel-accurate annotated images. In this architecture, we leverage both diffusion models and GANs to generate realistic paired images and their binary segmentation masks (image-mask pairs). We design a dual diffusion model setup for image and mask synthesis, connected through skip connections and cross-attention mechanisms to enhance the coherence between the generated images and their corresponding generated masks. To improve the quality of the synthetic data, we apply super-resolution techniques to enhance the resolution of generated image-mask pairs. As a use case, we evaluate the proposed methodology in the agricultural domain, using it for a downstream wheat head semantic segmentation task. Specifically, we externally evaluate the performance of a segmentation model that we train only on our generated data. We then use the final domain adaptation technique described in \cite{Najafian2022SemiSelfSupervisedLF} to improve the performance of the trained model.\par
This work presents a deep model architecture to synthesize annotated agricultural data; however, our method is applicable in any domain requiring semantic masks. We demonstrate its efficacy by generating high-quality image-mask pairs, advancing synthetic data generation techniques, and mitigating data scarcity in precision agriculture. Additionally, our results show enhanced realism of synthetic data and promising results for improving performance in downstream semantic segmentation tasks.%
\section{Related Work}%
Image segmentation has been employed to develop various agricultural automation tools~\cite{Nayak2024ImprovedDO,Gao2022AutomaticTD}. These tools can assist farmers in making informed decisions regarding crop coverage, disease identification, and weed control~\cite{Gao2022AutomaticTD}.ver, most such approaches rely on supervised methods that are constrained by the limited availability of annotated data~\cite{LUO2024172}.\par 
%
\noindent\textbf{Semantic Segmentation.} A semantic segmentation task can be modeled by a function $f:\mathbb{R}^{H \times W \times C} \rightarrow \mathbb{R}^{H \times W}$. The goal is to assign a discrete label to each image pixel. Recent state-of-the-art methods, such as~\cite{segmentation_1, segmentation_2, seg3}, have explored supervised segmentation using Convolutional Neural Networks. However, these algorithms require human-annotated data to learn, which limits their applicability in tasks for which large-scale annotated datasets are not available. Manual annotation of large-scale datasets is often impractical due to its high cost and time-consuming nature. In contrast, synthesizing computationally annotated data offers an effective alternative, enabling the development of robust deep learning models without the drawbacks associated with manual annotation~\cite{chen2021understanding}.\par
\noindent\textbf{Synthetic Data.}
Several studies have used 3D rendering software to generate synthetic images along with pixel-accurate labels~\cite{rendering1, rendering2, rendering3}; however, these images can be easily identified by humans as synthetic due to their visual characteristics, introducing a domain gap with real data. Other works, such as~\cite{Najafian2022SemiSelfSupervisedLF, alireza_domain}, have employed copy-paste methods to create synthetic data, and apply domain adaptation techniques to address the domain gap. \par
To address data annotation drawbacks, semi-supervised and self-supervised learning approaches~\cite{Goodfellow2022GenerativeAN,alireza_domain} have been utilized to train deep learning models with minimal or no data annotation. Conversely, generative approaches~\cite{latentdiffusion} mitigate data annotation limitations by creating synthetic yet realistic data, which is often computationally annotated for downstream tasks. This synthetic data is then used in developing deep learning models across semi-, self-, and fully-supervised learning paradigms. \par 
Several studies have explored the effectiveness of semi-supervised and self-supervised learning approaches in training deep learning models with computationally annotated or synthesized data. Ghanbari et al.~\cite{alireza_domain} studied the importance of domain adaptation in developing deep learning models, using solely computationally annotated synthetic data generated from only three manually annotated image samples. Myers et al.~\cite{Myers2024EfficientWH} utilized computationally annotated images, obtained using the cut-and-paste approach developed by~\cite{Najafian2022SemiSelfSupervisedLF} to train a modified CycleGAN which converts unrealistic synthetic images to images which appear realistic.\par
\noindent\textbf{Generative Adversarial Networks.}
Numerous studies have investigated the application of Generative Adversarial Network (GAN) models for data generation~\cite{Han2018GANbasedSB, Zhang2021StyleSwinTG}. A GAN consists of two components: a generator $G$ that creates samples from a random prior $z$ in a single pass, and a discriminator $D$ that learns to distinguish between real and generated samples. The training of these networks is formulated as a min-max process, where both $G$ and $D$ attempt to deceive each other. Upon completion of the training, $G$ can be used to generate new samples.\par 
Previous works using GANs have explored the utility of applying the forward diffusion process to images before classifying them by the discriminator $D$. DiffusionGAN~\cite{wang2023diffusiongantraininggansdiffusion} finds that using a standard GAN generator and conditioning $D$ on $t$, where $t$ represents the time steps in the Diffusion process, can stabilize training and yield superior results in data generation.\par
\noindent\textbf{Denoising Diffusion Probabilistic Models.}
DDPMs are latent variable models initially applied to unconditional data generation tasks. DDPMs have been shown to produce superior results, such as generating more diverse images, when compared to other generative approaches, including GANs and Variational Autoencoders~\cite{ho2020denoisingdiffusionprobabilisticmodels}. Furthermore, practical evidence indicates that DDPMs are less prone to mode collapse during training~\cite{rombach2022highresolutionimagesynthesislatent, wang2023diffusiongan}. Due to their flexibility and strengths, DDPMs have been expanded from unconditional image generation models to conditional models for use in many tasks such as in-painting~\cite{lugmayr2022repaintinpaintingusingdenoising}, super-resolution~\cite{moser2024diffusionmodelsimagesuperresolution}, and text to image generation~\cite{latentdiffusion}.
In this class of generative models, a neural network is trained in a self-supervised manner by iteratively denoising input images corrupted by Gaussian noise~\cite{ho2020denoisingdiffusionprobabilisticmodels}. Diffusion involves two main phases: the forward and the reverse processes.\par 
In the forward process, a data point $x_0$ from the dataset is gradually corrupted by adding Gaussian noise in $T$ successive steps. Forward diffusion follows a Markov chain process~\cite{Salimans2014MarkovCM}. This stochastic process is formally defined as follows: 
\begin{equation}
    x_t = \sqrt{1-\beta_t}x_{t-1} + \sqrt{\beta_t}\epsilon 
\end{equation}
where $\epsilon \sim \mathcal{N}(0, I)$ represents the sampled noise, and $\beta_t > 0$ is a value that specifies the noise variance at each timestep. Additionally, $t \le T$, where $T\in\mathbb{N}$ is set to be a large number, commonly $1000$, to increase the stability of the model optimization process and improve the quality of generated images. At sufficiently large values of $t$, $x_t$ converges to pure Gaussian noise.\par 
The reverse process aims to recover the original image by denoising $x_t$ to $x_{t-1}$ in each step. We initiate the process with $t=T$ and continue back to $x_0$ by performing $T$ iterative backward processes. 
To train the network, Gaussian noise of a random magnitude is added to an image at each iteration. The network is then tasked to predict the added noise, and the Mean Squared Error loss is calculated between the predicted and actual noise. The model, $f$, is conditioned on time step $t\in \mathbb{N}$, and is a function $f: \mathbb{R}^{H\times W\times C}\times\mathbb{N} \rightarrow \mathbb{R}^{H\times W\times C}$.\par
\noindent\textbf{Conditioning Image Diffusion Models.}
Numerous methods have been proposed to add conditioning to DDPMs for guided sample generation. ControlNet~\cite{controlnet} introduces conditioning to a pretrained (Latent) Diffusion Model through zero-initialized convolution layers, called zero-convolutions. This method creates an image-to-image model, which generates data following conditioning priors such as normal maps, depth maps, and canny edges. InstanceDiffusion~\cite{instancediffusion} allows for instance-level conditioning, where bounding boxes, masks, points, scribbles, and text can be used to condition each object instance in an image. This is achieved through the use of novel ScaleU and UniFusion blocks. 


Unlike the discussed generative studies, which focus solely on data generation without any annotation, this work aims to develop a method that leverages Diffusion and optimizes it in a GAN-style framework for generating images alongside their pixel-accurate segmentation masks. Regarding paired image-mask generation, Toker et al.~\cite{satsynth} developed SatSynth by training a single diffusion model $G$ on concatenated image-semantic mask pairs and applying generated data as a form of data augmentation in developing deep learning models for the task of satellite image segmentation. Han et al.~\cite{medgen} proposed MedGen3D for generating 3D volumetric pairs of brain MRI images/masks through a two-phase 2D image and mask sequence generation. DatasetGAN~\cite{styleganlabbeled} used a StyleGAN model as its backbone and Style interpreter to generate images and predict their corresponding semantic masks by fine-tuning the model for image generation and adding a new label generation branch for pixel-wise mask classification.\par 

Our work differs from the previously discussed studies. We propose an end-to-end generative architecture that combines Diffusion and GAN techniques for the simultaneous generation of 2D images and their corresponding masks. This approach is designed to handle images that contain densely packed and small objects of interest, such as those found in precision agriculture, and necessitates pixel-accurate, fine-grained binary segmentation masks. We design a novel architecture that allows for clear separation between image and mask features and improves the quality of the generated samples. All methods reviewed, other than SatSynth, either require pretrained weights, lack end-to-end functionality, or perform conditioning and require human intervention to initiate data generation. SatSynth aims at a similar objective; however, our method outperforms this concatenation approach in downstream precision agriculture tasks, as demonstrated in this study.\par

%
\section{Data and Methodology}
In this work, we propose a novel approach, \methodname, for generating computationally annotated synthetic data, particularly agricultural data.\par
\begin{figure}[!tbph]
    \centering
    \includegraphics[width=0.7\linewidth]{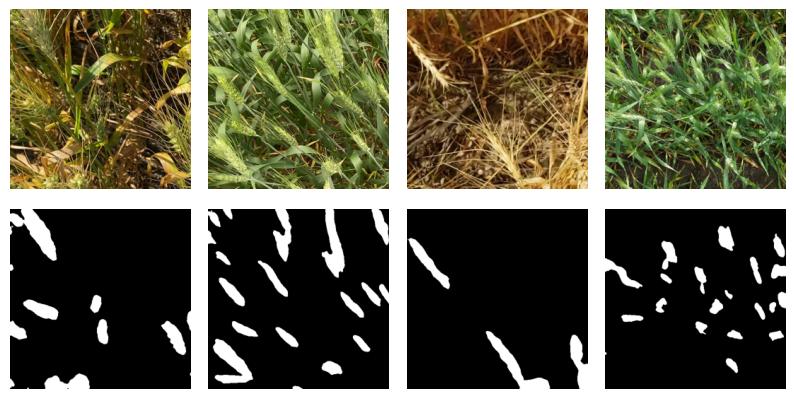}
    \caption{Data samples used for training, and their corresponding pseudo labels.}
    \label{fig:samples}
\end{figure}

\noindent\textbf{Data.}\label{Data}
We created a large-scale dataset of wheat head images from two sources. First, we extracted frames from a video of a wheat field captured on a Samsung camera, obtaining $10,183$ unlabeled samples from a single domain. Additionally, we used drone-captured videos to extract $16,148$ images of wheat fields covering different domains. All of these images were compiled into one dataset. The pseudo-labels for this dataset were obtained using the model developed by Najafian et al.~\cite{Najafian2022SemiSelfSupervisedLF}.
All images were segmented at resolution $1024 \times 1024$. Figure~\ref{fig:samples} shows examples of these images and their corresponding pseudo-labels. \par 
\noindent\textbf{Methodology.}
We designed a unique CNN-based architecture and discuss the effectiveness of combining GANs with diffusion models to enhance the realism of generated samples. We trained our method end-to-end to generate images paired with their corresponding semantic masks. Given the set $\Delta \subset \mathbb{R}^{H\times W\times C}$ of all images containing wheat heads and the set $\Theta \subset \mathbb{R}^{H\times W\times 1}$ of all wheat head segmentation masks, we aim to model the function $f: z \in \mathbb{R}^{H\times W\times (C+1)} \rightarrow \Psi$ such that $z \sim \mathcal{N}(0, 1)$, a standard normal distribution, and $\Psi \subset \Delta \times \Theta$ contains each image matched with its corresponding segmentation mask.\par
The model architecture consists of two parts, a diffusion model, $G$, which learns the joint distribution of images, $x$, and their segmentation masks, $y$, modeling $p(x, y)$, followed optionally by a discriminator, $D$, which learns the distribution of real images. $D$ is conditioned on the timestep $t$, and is a function defined as $f: \mathbb{N}\times\mathbb{R}^{H\times W\times (C+1)}\rightarrow \{0,1\}$.\par 
\begin{figure}
    \centering
    \includegraphics[width=0.9\linewidth]{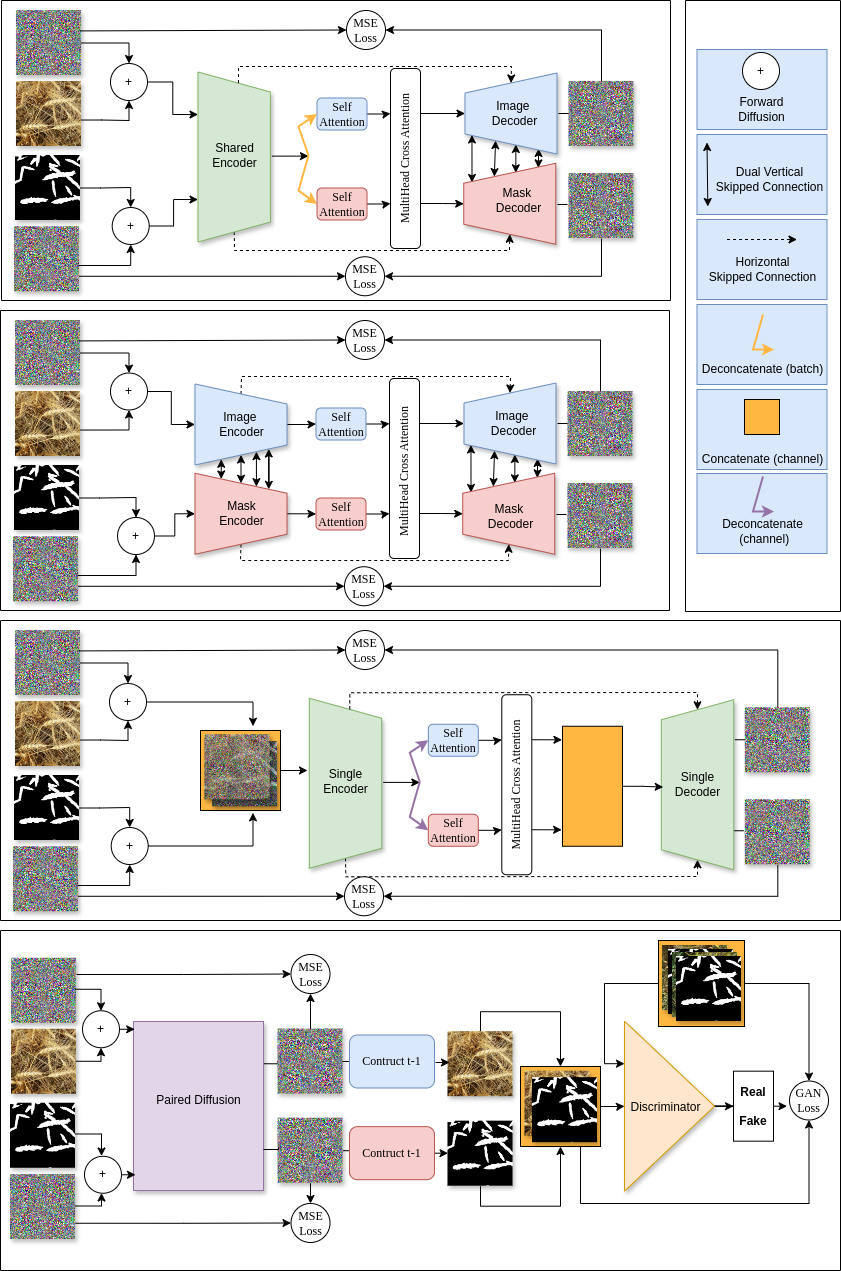}
    \caption{Our proposed pipeline, displayed from top to bottom: \textit{SharedEncoder} approach, \textit{TwoEncoder} approach, \textit{Concat} approach, and GAN approach. In the GAN approach, Paired Diffusion can be implemented using either \textit{SharedEncoder}, \textit{TwoEncoder}, or \textit{Concat}.
    }
    \label{fig:pipeline}
\end{figure}%
We experimentally evaluated the proposed method, \methodname, using three distinct variations of our architecture (Figure~\ref{fig:pipeline}): \textit{TwoEncoder}, \textit{SharedEncoder}, and \textit{Concat}. We utilized two distinct diffusion models, $G_x$ for images and $G_y$ for segmentation masks, both based on the U-Net architecture. These models communicate via skip connections after each down-sampling block, up-sampling block, and through cross-attention in the bottleneck layer. We compare this method to a similar \textit{SharedEncoder} method. We propagate both the image $x$ and mask $y$ independently through a shared encoder and bottleneck $E$, to compute $x_l = E(x)$ and $y_l = E(y)$. Finally, $x_l$ and $y_l$ are propagated separately through two unique decoders $D_x$ and $D_y$, respectively, which communicate via skip connections. In both \textit{SharedEncoder} and \textit{TwoEncoder} approaches, $x_l$ and $y_l$ undergo self- and cross-attention, which is a common way of integrating conditioning to diffusion models. Additionally, as proposed in~\cite{satsynth}, which is, to the best of our knowledge, the only other work that uses diffusion to unconditionally generate datasets, we design a model with a single encoder and a single decoder. However, in these experiments, we train the model on the concatenated images and masks in the channel dimension. We refer to this approach as the \textit{Concat} method.\par 
Our proposed pipeline, along with its architectural variations, is illustrated in Figure~\ref{fig:pipeline}. To train our models, we used real images and their corresponding pseudo-labels (see section~\ref{Data}). The implementation details and the code for~\methodname~can be found at \url{https://github.com/VisionResearchLab/SynthSet}.\par
\noindent\textbf{Discriminator.}
To further align the generated images to the real domain, we add a discriminator, $D$, to our end-to-end pipeline. DiffusionGAN~\cite{wang2023diffusiongantraininggansdiffusion} explores the use of the forward diffusion chain for GAN training. They found that this results in stable training and exceptional image generation in GAN models. As proposed in~\cite{wang2023diffusiongantraininggansdiffusion}, we employed a time-dependent discriminator, which receives both real and generated images at timestep $t$ of the diffusion process. To accomplish this, after $G$ predicts the noise which was added to $x_{t}$ and $y_{t}$, we sample $x_{t-1}$ and $y_{t-1}$. These are given to the discriminator, concatenated, along with real image-mask pairs at time $t-1$. 
\par 
When training with a discriminator, we apply a schedule to the maximum timestep sampled in each epoch, similar to the approach in~\cite{wang2023diffusiongantraininggansdiffusion}. This allows the discriminator to learn first from simpler tasks, specifically lower timesteps. At low values of $t$, the content of images can be clearly understood; however, at higher values of $t$, images converge to noise. By initiating training at low timesteps, the discriminator can gradually learn the complexities present in higher timesteps. The schedule which we use is defined as follows:
\\\\ 
\centerline{$t_{max} = min[T, \sigma\times (s/\alpha)+i]$} 
\\\\
where $t_{max}$ represents the maximum step to sample for the forward pass, $T$ denotes the maximum timestep of the schedule, $\sigma$ is the step size, $s$ is the current step, $\alpha$ is the number of epochs it takes to step, and $i$ is the initial maximum. To further assist with training, we priority sampled higher values of $t$ until epoch 500, so the network could focus on learning the newly exposed time steps.\par
%
\noindent\textbf{Skip Connections.}
Skip connections involve using the output of a block $b$ in a network as part of the input of block $c$ in the network, where $b$ and $c$ are not subsequent blocks. Formally, block $c$ is computed as $c(x, b(x))$ rather than $c(x)$. Skip connections are often used in neural networks to propagate features across residual blocks, and in U-Nets to transfer high-level features to the decoder~\cite{unet}.
In our \textit{TwoEncoder} and \textit{SharedEncoder} method, the skip connections across separate encoders and decoders are the primary source of communication between mask generation and image generation modules.
With these we model $p(x,y)$, as opposed to $p(x \mid y_t)$ and $p(y \mid x_t)$.
To evaluate the superior configuration for skip connections, we compare concatenating the features (1) directly, (2) using zero-initialized convolutions (ZeroConv) as proposed in ControlNet~\cite{LUO2024172}, and (3) using a variation of the ScaleU block proposed in InstanceDiffusion~\cite{instancediffusion}, modified to scale and accept an arbitrary number of skip connections.\par

\begin{figure}[!th]
    \centering
    \includegraphics[width=\linewidth]{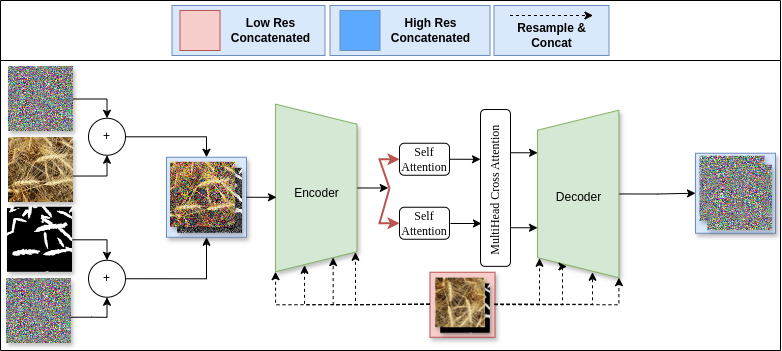}
    \caption{Super-resolution pipeline. High-resolution images are of size $256\times256$, while the low-resolutions are $128\times128$.}
    \label{fig:super_res}
\end{figure}

%
\noindent\textbf{Super-resolution.}
DDPMs are often applied to images of resolution $128\times 128$. While these algorithms can be easily extended to higher resolutions, the computational overhead grows rapidly, especially with the inclusion of attention modules.
%
As such, we trained a super-resolution diffusion model, which is an image-to-image DDPM that generates high-resolution ($256\times 256$) images from the generated low-resolution image-mask pairs ($128\times 128$). We utilized bilinear resampling for images and nearest neighbor for masks. Our super-resolution network was trained using the same dataset as for all experiments (see Section \ref{Data}). To condition the model, we concatenated the low-resolution image-mask pairs to the tensor before each down-sampling and up-sampling block. This method can be seen in~Figure~\ref{fig:super_res}.\par

%

Generating a dataset with 5000 images requires up to two hours on a single A100 GPU using our largest models, presenting a bottleneck for model development. DDPMs excel at generating realistic and diverse images; however, the time required for generating large datasets can become an issue, especially when compared to other methods of generating synthetic data, such as copy-paste and GAN-based methods. To address this, we used the Denoising Diffusion Implicit Model (DDIM)~\cite{ddim} inference methodology to accelerate data generation by reducing the number of timesteps.\par 
%
%
%
\noindent\textbf{Training the Proposed Architectures.}
We trained our networks on either a single Nvidia $A6000$ or a single Nvidia $A100$ GPU, with a batch size of $64$. We sampled with $T=1000$ utilizing a linear scheduler as described in~\cite{ho2020denoisingdiffusionprobabilisticmodels}. We employed the Adam optimizer~\cite{Kingma2014AdamAM}, with $lr=0.00021$. We used Mean Square Error (MSE) as the objective function for the diffusion models, and a standard adversarial loss using Binary Cross-Entropy for the discriminator. When applicable, the weight of the loss being propagated from the discriminator to the generator was a quarter that of the diffusion MSE. 
We used a random cropping augmentation to extract a patch of $512 \times 512$ pixels, which was then resized to $128 \times 128$. Subsequently, as described, our downstream super-resolution model upsamples this to $256 \times 256$.
We trained for $1,500$ epochs, which is $231,750$ optimization steps. We applied random horizontal and vertical flips as augmentations. To resize our images, we used bilinear resampling, and for the masks, we used nearest neighbor resampling. To monitor for overfitting and to effectively evaluate super-resolution, our data was split into train and validation sets with a ratio of 80/20. Images were scaled to the intensity range $\left[0, 1\right]$ before training.\par
%
%
\noindent\textbf{Training the Downstream Segmentation Models.}
We evaluated the utility of our generated images by using them in a segmentation task. To evaluate each approach within \methodname, we trained a segmentation U-Net model from scratch using $5000$ synthetic samples generated by each method.
As proposed by Najafian et al.~\cite{Najafian2022SemiSelfSupervisedLF} for the U-Net encoder, we use an EfficientNet-B4~\cite{en} model and an inverted version for the decoder. The models were trained with the SGD~\cite{ruder2016overview} optimizer, $0.01$ learning rate, and for $50$ epochs. We used Dice and BinaryCrossEntropy losses combined in training all the downstream segmentation task models. Note that we consider the method for generating data which yields the best performance in this downstream task as being the best method for generating data.
\section{Results}
\begin{figure}[!th]
    \centering
    \includegraphics[width=0.6\linewidth]{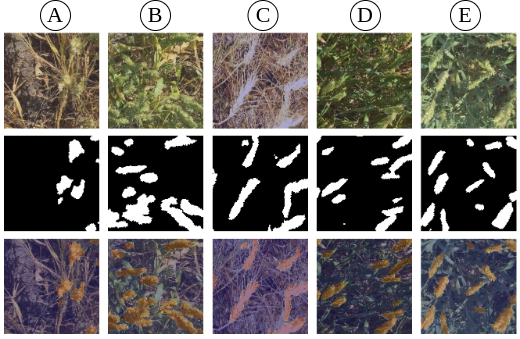}
    \caption{Images generated by different variations of our model architecture. The top row displays the images, the center row shows the masks, and the bottom row showcases the images overlaid by their corresponding masks. The columns display \methodname~variations in the following order: (A) \textit{Concat}, (B) \textit{Concat} with Discriminator, (C) \textit{TwoEncoder}, (D) \textit{TwoEncoder} with Discriminator, and (E) \textit{SharedEncoder}. Although each method generates realistic images, adding the discriminator to each variation further enhances the depth and realism of the generated wheat images.
    }
    \label{fig:gen-samples}
\end{figure}

\begin{figure}[!th]
    \centering
    \includegraphics[width=0.6\linewidth]{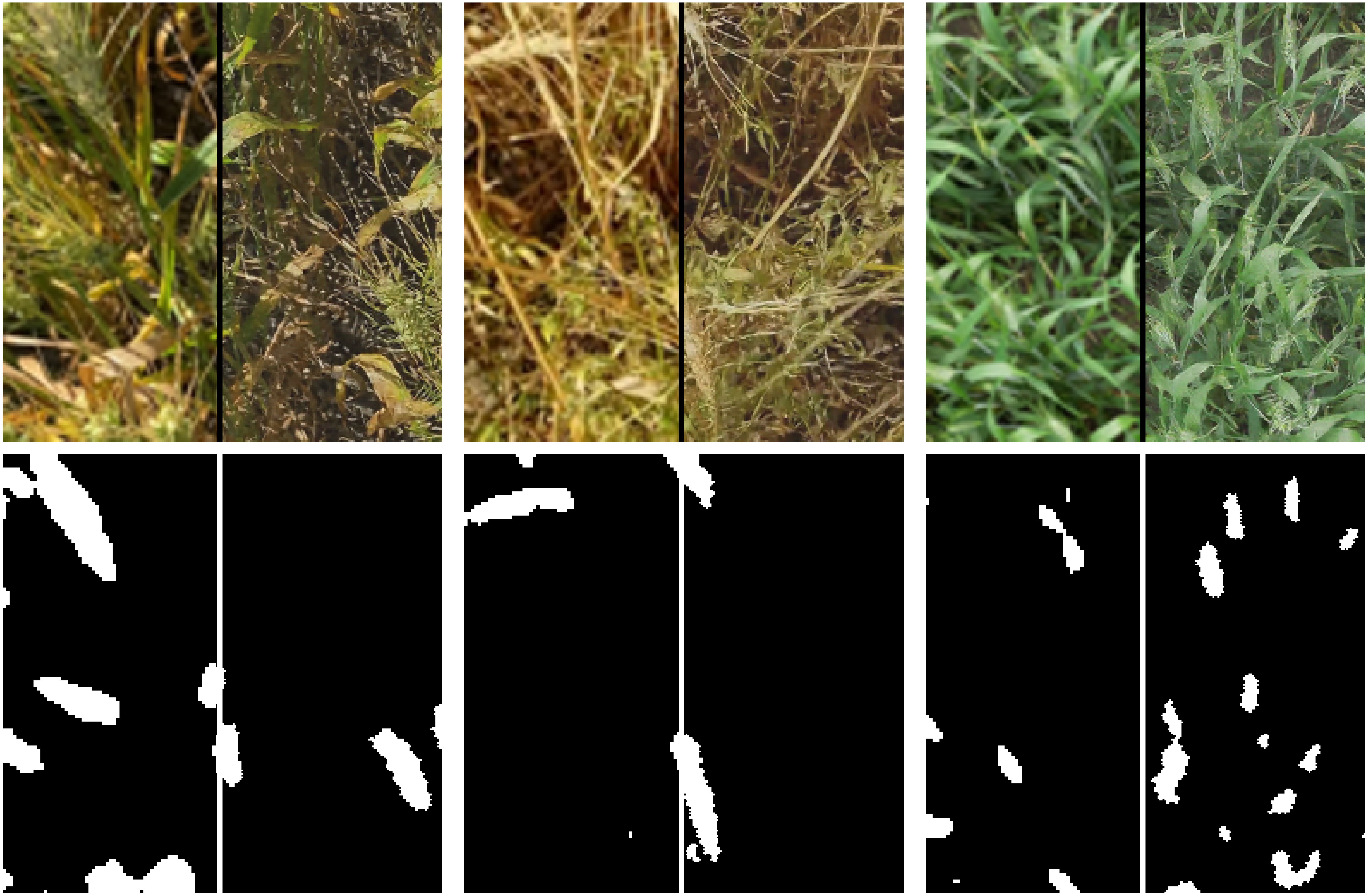}
    \caption{Images and their masks split left and right. Left is bilinear upsampling for images and the nearest neighbor for masks. Right is the result of our super-resolution diffusion model.}
    \label{fig:super_res_samples}
\end{figure}

Figure~\ref{fig:gen-samples} shows examples of generated images and their corresponding masks. We observe in numerous generated images, that models trained with a discriminator generate more realistic-looking images, particularly, these models generate images with more depth. Furthermore, the high-resolution image-mask pairs generated by our super-resolution model are illustrated in Figure~\ref{fig:super_res_samples}.\par
As described in the Skip Connections section, we run experiments with three methods of skip connection management. Qualitatively, we observe that the use of the ZeroConv results in poor convergence of the mask branches, and the absence of skip connection moderation (direct concatenation) results in poor image quality. The utility of the ScaleU is evident and adds minimal overhead to the network parameter, thus, we use this block in our method.\par
\noindent\textbf{External Evaluation.}
We evaluated the performance of our various segmentation models using a hand-annotated test set of $365$ images from the Global Wheat Head Dataset (GWHD)~\cite{david2021globalwheatheaddataset} at a resolution of $1024\times 1024$. We crop each image in the GWHD to $512\times 512$ pixels and then resize to $256 \times 256$. Consequently, each image in the original GWHD yields four images in our test set, resulting in $1460$ test images.\par
%
\noindent\textbf{Weight Selection.} 
Evaluating diffusion models during training is challenging, particularly in balancing image quality with variability during sampling. We considered three ways for selecting model weights: (1) best validation loss; (2) final epoch ($1500$); and (3) best mean Jensen Shannon Divergence (divergence) between the training RGB distribution, and the RGB distribution of $64$ generated images. The rationale behind monitoring divergence is that wheat field images, dominated by vegetation, exhibit distinct color domains. Thus, we hypothesized that weights generating RGB images with a distribution similar to that of the train set would produce realistic and diverse samples. Using each \methodname~approach, we compared the weight selection methods to determine which model achieves superior downstream performance. We observed that using the lowest mean divergence yields the best results for some methods; however, other methods performed better with the final epoch. Notably, selecting weights based on the lowest validation loss obtained suboptimal outcomes. We conclude that to generate the best datasets, it is crucial to consider metrics such as divergence, loss, and iteration count. All other experiments described in the following sections use the weights that yield the best downstream performance.\par
\begin{figure}[!b]
    \centering
    \includegraphics[width=0.6\linewidth]{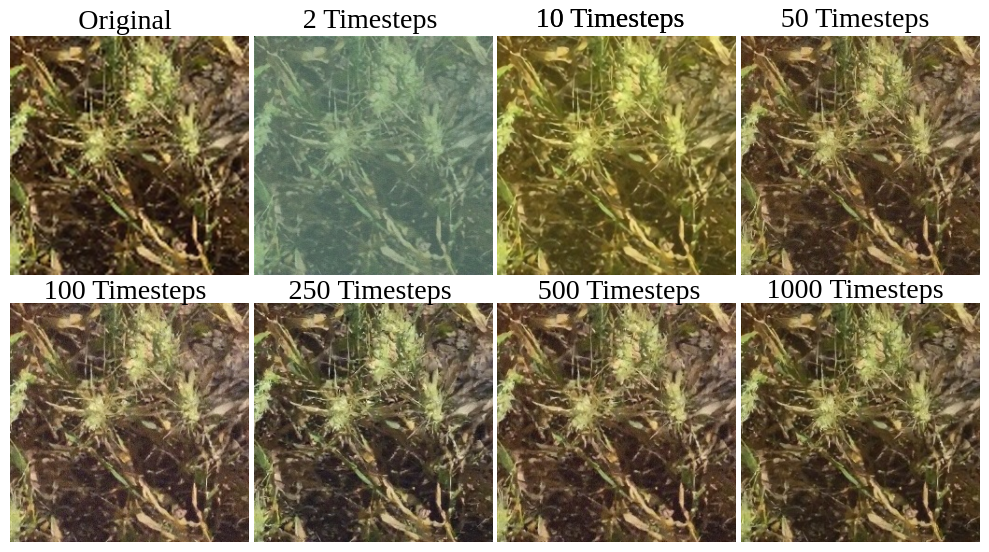}
    \caption{Results from using different DDIM inference steps. The original image at the top left has been resized using bicubic interpolation. Downstream tasks using images super-resolved with greater than or equal to 100 DDIM steps perform very similarly.}
    \label{fig:ddim-steps}
\end{figure}

 
%
\noindent\textbf{DDIM Inference.}
Image generation and super-resolution using DDPMs is a time-consuming task and requires iterating over 1000 timesteps twice during reverse diffusion. It can take up to two hours on a single $A100$ GPU to generate 5000 samples, with the super-resolution step accounting for the majority of the time. To speed up the generation process, we explored the utility of using the inference methodology described in~\cite{ddim} for super-resolution. We maintained 1000 DDPM inference steps for image generation to ensure we obtained the highest quality samples\cite{ddim}. On two batches of data, we super-resolved images generated by the \textit{Concat} version of \methodname, using weights obtained at the final epoch. We evaluated the downstream segmentation task using 1000, 500, 250, and 100 DDPM steps. We then used the same steps but using DDIM. The detailed results on the downstream task with two sets of 5000 image-mask pairs can be seen in appendix~\ref{app:ddim}, tables~\ref{tab: ddim-results} and~\ref{tab: ddim-results2}. In both experiments, we reached the same conclusion. The 1000-step DDPM performs best; however, reducing DDPM steps decreases performance. As shown in appendix~\ref{app:ddim}, when using DDIM inference with only 100 steps, we experienced a negligible loss in performance in the downstream task; however, the fine-tuning resulted in up to a $2\%$ performance improvement. As such, we determined that the $10\times$ speed-up justifies a small reduction in accuracy. We moved forward with 100 DDIM steps for super-resolution in all further experiments. The impact of using different DDIM steps for super-resolution is illustrated in Figure~\ref{fig:ddim-steps}.\par
%
%
\noindent\textbf{Segmentation}
We assessed the utility of data generated by different model architectures in our pipeline, \methodname, for developing segmentation models and compared their results with those of the model trained on data generated using the copy-paste approach by Najafian et al. (see Figure~\ref{fig:copypaste} in Appendix~\ref{app:samples}). The models trained on our synthetic data outperformed copy-paste (see Table~\ref{tab: final}), the best results achieved by the \textit{TwoEncoder} \methodname~without a discriminator. The method described in Najafian et al. contains four steps, in which step one consists of training on copy-paste data. Steps two and three address the domain shift between synthetic and real images. Then, step four addresses the domain shift between different wheat domains and growth stages, using images from 18 domains in the GWHD. In our implementation of this pipeline, we reduced the number of synthetic samples used to 5000 and the resolution to $256\times 256$ to match our experiments. Unlike Najafian et al.~\cite{Najafian2022SemiSelfSupervisedLF}, we fine-tuned on images from different growth stages of all 18 domains in GWHD immediately after training on synthetic images. Since the images generated by our models are realistic and diverse, we omitted steps two and three, because bridging the domain gap between synthetic and real images is no longer a concern. Finally, we evaluated the trained model on the GWHD test set, outperforming the state-of-the-art methodology. Dice and IoU scores for each of our methods are presented in Table~\ref{tab: final}. Note that the GWHD dataset was not used for model training, and is, therefore, an external test set. 

\begin{table}[!tbph]
\centering
\caption{Results on the downstream segmentation task, using various synthetic data methods, and fine-tuning methods.}\label{tab: final}
\scriptsize
\renewcommand{\arraystretch}{1.2}
\begin{tabular}{|l|cc|cc|cc|}\hline  
&\multicolumn{2}{p{2.3cm}|}{\centering \makecell{Synthetic \\ Data}} 
&\multicolumn{2}{p{2.8cm}|}{\centering \makecell{Synthetic + \\ GWHD Fine-tune}} 
&\multicolumn{2}{p{2.8cm}|}{\centering \makecell{Synthetic + \\ All Fine-tuning}} \\\hline
\textbf{Method Flavor} &\textbf{Dice} &\textbf{IoU} &\textbf{Dice} &\textbf{IoU} &\textbf{Dice} &\textbf{IoU} \\\hline
\textit{Concat} &0.7254 &0.6033 &0.758 &0.6409 &0.6381 &0.5046 \\\hline
\textit{Concat} w. Discrim &0.7225 &0.6013 &0.7661 &0.6548 &0.7154 &0.5937 \\\hline
 \textit{SharedEncoder} &0.7491 &0.6333 &0.7712 &0.6594 &0.71534 &0.5937 \\\hline
 \textit{TwoEncoder} &\textbf{0.7611} &\textbf{0.6496} &\textbf{0.7927} &\textbf{0.6908} &0.7044 &0.581 \\\hline
\textit{TwoEncoder} w. Discrim &0.7221 &0.6004 &0.7741 &0.6642 &0.6773 &0.5495 \\\hline
Copy-Paste (Najafian et al.) &0.4172 &0.2944 &0.5534 &0.4141 &0.6293 &0.4962 \\
\hline
\end{tabular}
\end{table}

As evidenced by our segmentation results, each version of our method, \methodname, yields realistic annotated data and can be used for downstream segmentation. In particular, \textit{TwoEncoder} \methodname~without a discriminator shows the best results, $2\%$ ahead of the second place \textit{SharedEncoder} variant. This surpasses the previous method for unconditional annotated data generation, SatSynth\cite{satsynth}.
\section{Discussion}
In this paper, we proposed \methodname~for synthesizing realistic and computationally annotated datasets for semantic segmentation tasks. \methodname, while training only with synthetic data, surpassed the state-of-the-art technique for data synthesis used in wheat head segmentation when evaluated on a diverse external dataset.\par
\par
Though \textit{TwoEncoder} without a discriminator outperforms the version with a discriminator, we have observed that images generated from models with a discriminator are visually superior; however, discriminator training reduces variability, leading to the lower downstream performance.\par
Semantic segmentation is inherently bottlenecked by the size and diversity of datasets. Using our method to generate new datasets or as an augmentation technique can improve segmentation performance, mitigating the data bottleneck. We suggest that future research explores the utility of data computationally generated by our method in domains outside precision agriculture.\par
One limitation of this work is the impact of weight selection on performance in downstream tasks. As we have shown, different epochs can yield varying results, and one cannot rely solely on the lowest validation loss or the final epoch to select weights. This is due to the trade-off between variability and image quality. Detailed results of these experiments can be seen in appendix~\ref{app:weights}, Figure~\ref{fig:weight_select_fig}. 
The model \textit{Concat} performed best at the least divergence epoch, which was on epoch $865$. This is likely due to the diversity of the images generated, which is evidenced by PCA embeddings for this model at the three discussed epochs. These can be seen in appendix~\ref{app:weights}, Figure~\ref{fig:pca}. Future research could explore conditioning \methodname~on image embeddings to enforce dataset variability and reduce the impact of weight selection on performance.
\par
The time taken to generate samples is an obstacle to the applicability of methods such as our own. We evaluated the efficacy of utilizing the DDIM inference method on both image sampling and image super-resolution, finding that the number of timesteps for super-resolution can be reduced to 100. An alternative approach to overcome computational challenges is to use a latent DDPM. Conditioned latent diffusion is explored in depth by~\cite{rombach2022highresolutionimagesynthesislatent}. The use of these models results in faster training, allowing for higher resolutions and much faster sampling. We ran the experiments of our method in the latent space; however, we found that the generated masks lacked accurate alignment with their corresponding images. We suggest future research to explore the applicability of latent diffusion to our method.
\par
We applied the ScaleU block to manage skip connections, after experimenting with ZeroConv and direct concatenation. We observed that the use of the ZeroConv resulted in poor mask generation; direct concatenation also resulted in poor agreement of the image-mask pairs. The employment of the ScaleU block addresses these issues effectively. As stated in InstanceDiffusion~\cite{instancediffusion}, the ScaleU block enables the network to identify and learn the most informative skip connection features and allows the network to selectively diminish, amplify, or erase the influence of features. These results suggest that the information passed, and not passed, by skip connections is vital, and that the separation of information between image and mask generation may be the reason why \methodname~\textit{TwoEncoder} and \textit{SharedEncoder} yield superior samples to the concatenation method described in SatSynth\cite{satsynth}.
\par
\begin{figure}[!tbph]
    \centering
    \includegraphics[width=0.6\linewidth]{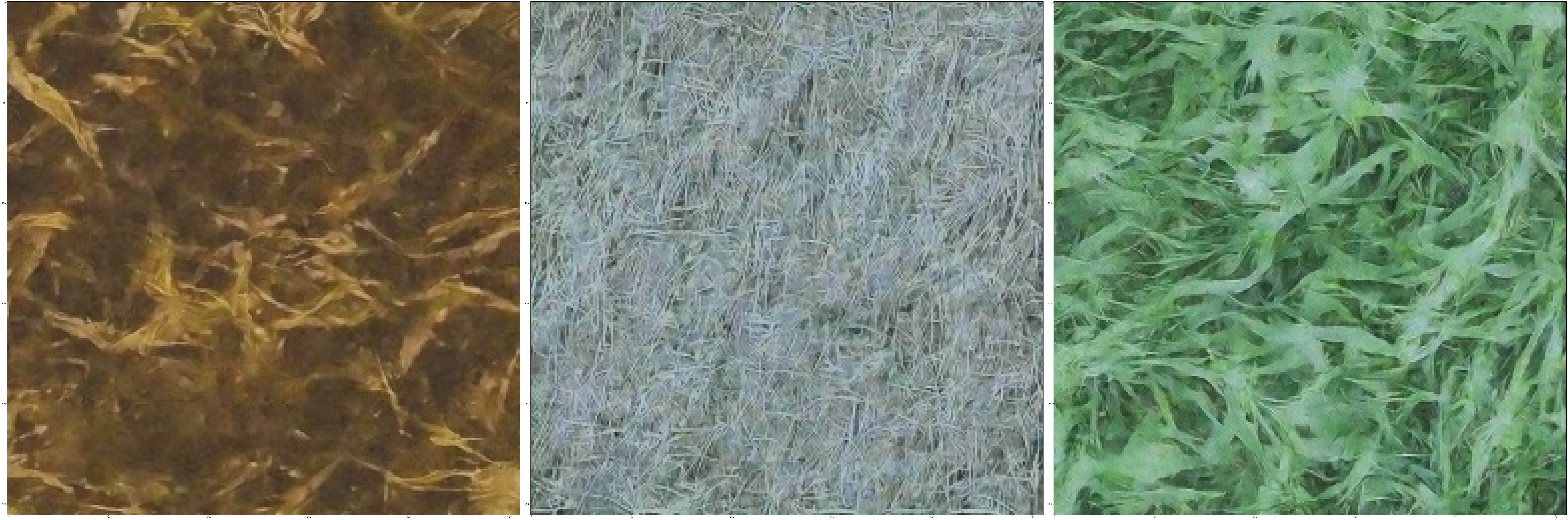}
    \caption{Examples of images generated by a model trained with a discriminator and without a timestep schedule. The results indicate that employing a discriminator during diffusion training necessitates the application of a timestep schedule.
    }
    \label{fig:nosched}
\end{figure}

When including a discriminator during training, we applied a schedule to the maximum timestep at each epoch. When no maximum timestep schedule is applied during training, images generated by these models are distorted, and unrealistic. This phenomenon can be seen in Figure~\ref{fig:nosched}. We conclude that to include a discriminator, this timestep schedule must be applied.
\par
\section{Conclusion}
In this paper, we proposed \methodname, a novel approach to the synthesis of image-mask pairs using a dual diffusion method. Our generated data improves upon the performance of the state-of-the-art method used for wheat head segmentation and yields competitive performance when used to train a segmentation model from scratch, allowing us to achieve a $76\%$ dice on the Global Wheat Head test set without training on any real images, and $79\%$ with a small fine-tuning set. 
Our proposed architecture sets a new benchmark for unconditional dataset-generating diffusion models.
While we assessed the utility of the proposed approach for synthesizing data for wheat head segmentation, the proposed methodology is not limited to this task and can be applied to various segmentation tasks, alleviating the need for large-scale manually annotated datasets.\par
\noindent\textbf{Acknowledgements}
We gratefully acknowledge the support of Google exploreCSR, which made this research possible. We thank the Natural Sciences and Engineering Research Council of Canada (RGPIN-2024-04966), Alberta Innovates, and the PURE program at the University of Calgary for their support.
%
%
\bibliographystyle{splncs04}
\bibliography{main}

\newpage

\appendix

\section{Model Weight Selection}\label{app:weights}
\begin{figure}[!tph]
    \centering
    \includegraphics[width=0.7\linewidth]{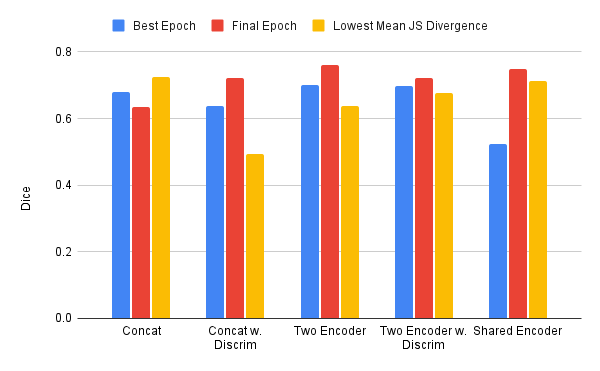}
    \caption{Results on the downstream task, without fine-tuning, comparing weight selection of best validation loss, the final epoch, and the epoch yielding the lowest mean JS divergence.}
    \label{fig:weight_select_fig}
\end{figure}
\vspace{30pt}
\begin{figure}[!tph]
    \centering
    \includegraphics[width=0.6\linewidth]{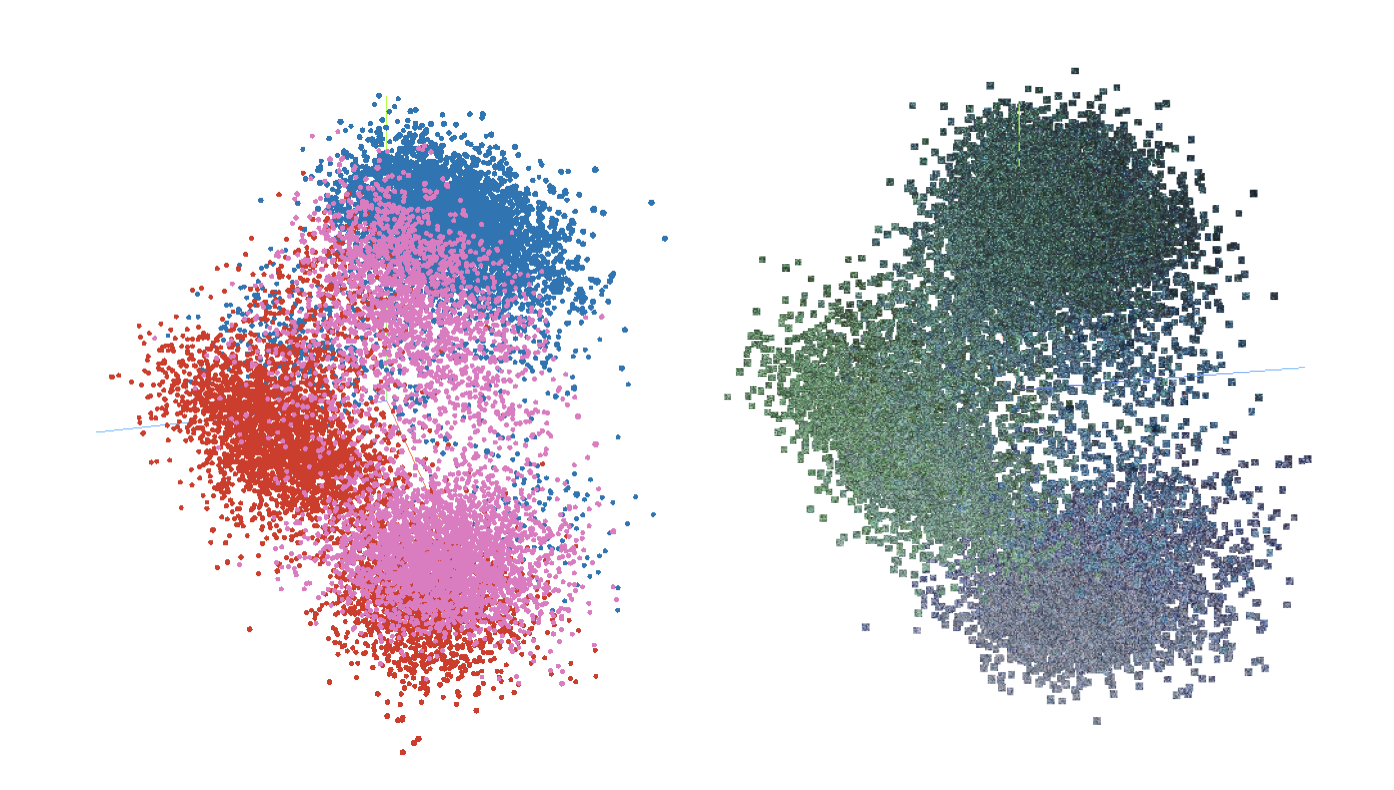}
    \caption{PCA of embeddings of images generated at different model selections. Pink points are the lowest mean JS Divergence, blue is the best epoch (lowest validation loss), and red is the final epoch (epoch 1500). The lowest divergence samples are much more spread, covering the domains that exist in both other methods, whereas the final and best epochs are almost mutually exclusive from each other regarding generated domains. To the right, the same embeddings are shown with low-resolution sprites, demonstrating the spread of domains.}
    \label{fig:pca}
\end{figure}

\section{Data Samples}\label{app:samples}

\begin{figure}[H]
    \centering
    \includegraphics[width=0.6\linewidth]{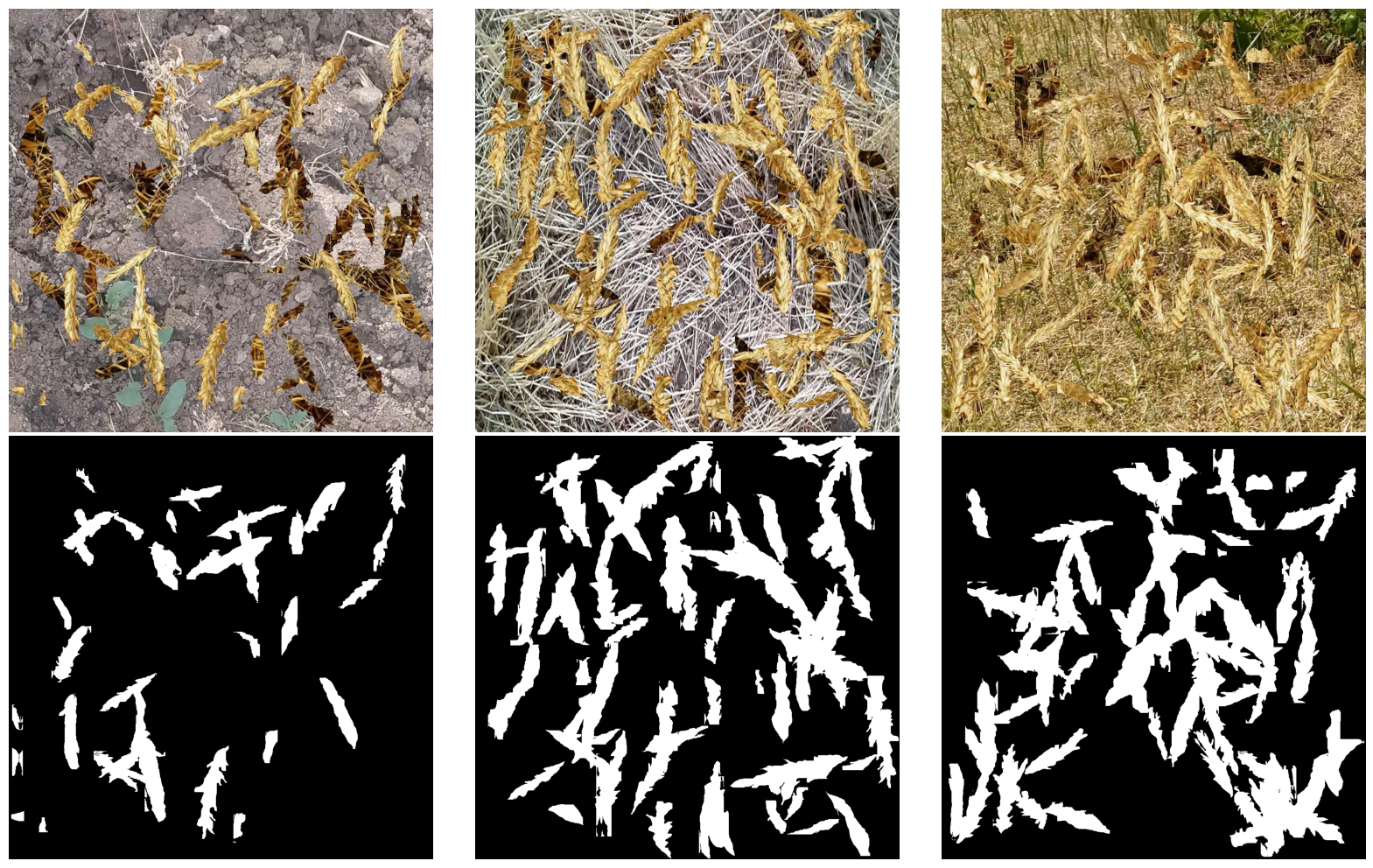}
    \caption{Synthetic data generated with copy-paste method described by Najafian et al.~\cite{Najafian2022SemiSelfSupervisedLF}.
    }
    \label{fig:copypaste}
\end{figure}

\section{DDIM Super Resolution Steps}\label{app:ddim}

\begin{table}[!htp]\centering
\caption{Downstream performance across different inference steps used for super-resolution on the same generated images. Image Batch 1.}\label{tab: ddim-results}
\scriptsize
\renewcommand{\arraystretch}{1.2}
\begin{tabular}{|l|cc|cc|}\hline
&\multicolumn{2}{p{2.3cm}|}{\centering \makecell{Generated \\ Only}} 
&\multicolumn{2}{p{2.8cm}|}{\centering \makecell{GWHD \\ Finetuned}} \\\hline
\textbf{SR Steps} &\textbf{Dice} &\textbf{IoU} &\textbf{Dice} &\textbf{IoU} \\\hline
1k ddpm &\textbf{0.6335} &\textbf{0.4995} &0.6466 &0.5143 \\\hline
500 ddpm &0.6237 &0.4882 &0.6559 &0.5238 \\\hline
250 ddpm &0.6221 &0.4863 &0.653 &0.5203 \\\hline
100 ddpm &0.5834 &0.4439 &0.6145 &0.4773 \\\hline
500 ddim &0.6258 &0.4913 &0.6521 &0.5187 \\\hline
250 ddim &0.6248 &0.4906 &0.6607 &0.5288 \\\hline
100 ddim &0.6269 &0.4931 &\textbf{0.6653} &\textbf{0.5353} \\\hline
\end{tabular}
\end{table}

\begin{table}[!htp]\centering
\caption{Downstream performance across different inference steps used for super-resolution on the same generated images. Image Batch 2.}\label{tab: ddim-results2}
\scriptsize
\renewcommand{\arraystretch}{1.2}
\begin{tabular}{|l|cc|cc|}\hline
&\multicolumn{2}{p{2.3cm}|}{\centering \makecell{Generated \\ Only}} 
&\multicolumn{2}{p{2.8cm}|}{\centering \makecell{GWHD \\ Finetuned}} \\\hline
\textbf{SR Steps} &\textbf{Dice} &\textbf{IoU} &\textbf{Dice} &\textbf{IoU} \\\hline
1k ddpm &\textbf{0.644} &\textbf{0.5115} &0.6408 &0.5079 \\\hline
500 ddpm &0.6305 &0.4956 &0.6641 &0.5336 \\\hline
250 ddpm &0.6296 &0.4951 &0.6527 &0.5198 \\\hline
100 ddpm &0.5952 &0.4572 &0.6191 &0.4838 \\\hline
500 ddim &0.6336 &0.5002 &0.666 &0.535 \\\hline
250 ddim &0.6383 &0.5057 &0.6671 &0.5363 \\\hline
100 ddim &0.6403 &0.508 &\textbf{0.6797} &\textbf{0.5504} \\\hline
\end{tabular}
\end{table}
\end{document}